# Estimation of forest height and biomass from open-access multi-sensor satellite imagery and GEDI Lidar data: high-resolution maps of metropolitan France

David Morin[1,*], Stéphane Mermoz[2], Florian Mouret[1,3], Milena Planells[1]

[1] CESBIO, Université de Toulouse, CNES/CNRS/INRAE/IRD/UPS
[2] Global Earth Observation (GlobEO)
[3] LBLGC, Université d'Orléans, USC1328 INRAE (Ecodiv)
[*] Correspondence: david.morin@univ-tlse3.fr

**Abstract**

Mapping forest resources and carbon is important for improving forest management and meeting the objectives of storing carbon and preserving the environment. Spaceborne remote sensing approaches have considerable potential to support forest height monitoring by providing repeated observations at high spatial resolution over large areas. This study uses a machine learning approach that was previously developed to produce local maps of forest parameters (basal area, height, diameter, etc.). The aim of this paper is to present the extension of the approach to much larger scales such as the French national coverage. We used the GEDI Lidar mission as reference height data, and the satellite images from Sentinel-1, Sentinel-2 and ALOS-2 PALSA-2 to estimate forest height and produce a map of France for the year 2020. The height map is then derived into volume and aboveground biomass (AGB) using allometric equations. The validation of the height map with local maps from ALS data shows an accuracy close to the state of the art, with a mean absolute error (MAE) of 4.3 m. Validation on inventory plots representative of French forests shows an MAE of 3.7 m for the height. Estimates are slightly better for coniferous than for broadleaved forests. Volume and AGB maps derived from height shows MAEs of 75 tons/ha and 93 m³/ha respectively. The results aggregated by sylvo-ecoregion and forest types (owner and species) are further improved, with MAEs of 23 tons/ha and 30 m³/ha. The precision of these maps allows to monitor forests locally, as well as helping to analyze forest resources and carbon on a territorial scale or on specific types of forests by combining the maps with geolocated information (administrative area, species, type of owner, protected areas, environmental conditions, etc.). Height, volume and AGB maps produced in this study are made freely available here: https://zenodo.org/doi/10.5281/zenodo.8071003.

# Table of contents

# List of figures

# 1. Introduction

Forests act as carbon sink by capturing carbon from the atmosphere through tree growth and act as carbon source through deforestation and wildfires. The monitoring of forest carbon stocks is crucial to assess the exchange of carbon between the earth's surface and the atmosphere and therefore to reduce the uncertainty in the global carbon budget. Forest height is a key component for forest carbon stocks estimation. Deriving forest aboveground biomass (AGB), closely related to carbon, from height relies on simple relations called allometric equations for a given tree species.

Spaceborne remote sensing approaches offer considerable potential in support of forest height monitoring as they provide repetitive observations over large areas. LiDAR data-based approaches have proven to be accurate enough to infer the canopy height and structure, and thus to map the forest AGB at a high spatial resolution (Dubayah *et al.*, 2020; Duncanson *et al.*, 2022). Since 2018, Global Ecosystem Dynamics Investigation (GEDI) is the first sensor providing high resolution laser ranging measures of forest height, canopy vertical structure, and surface elevation, from the International Space Station (ISS). Although sparse, lidar data from GEDI can be combined with spaceborne optical and radar data to provide continuous wall-to-wall maps of forest height. Among notable examples, (Potapov *et al.*, 2021) used Landsat time series and GEDI samples with a machine learning approach to produce a global forest canopy height map at 30m spatial resolution for the year 2019. (Lang *et al.*, 2022) used Sentinel-2 and GEDI in a convolutional neural network to produce another global forest canopy height map at 10m spatial resolution for the year 2020. On a local scale, (Morin *et al.*, 2022) combined Sentinel-2 optical sensor and Sentinel-1 and ALOS-2 PALSA2 radar sensors with GEDI to estimate forest canopy height with a machine learning approach, assessing the use of GEDI samples to replace field height measurements in model calibration. Finally, (Schwartz *et al.*, 2022, 2023) combined Sentinel-1 and Sentinel-2 in a deep learning U-Net model to first produce a local map of forest height, the method was then extended to produce a national map for France.

The processing chain used in (Morin *et al.*, 2022) has been previously developed and validated using local forest measurement datasets (Morin *et al.*, 2018, 2019; Morin, 2020). It allows to map several forest structure parameters: tree height, basal area, diameter at breast height (DBH), tree density, volume and aboveground biomass. The combined use of both optical and radar sensors, as well as spatial texture indices derived from images at 10m spatial resolution, is important to obtain good estimates of the different forest variables (Morin *et al.*, 2019). The use of these satellite image features and geolocated measurement data with a Random Forest (RF) or a Support Vector Regression (SVR) showed very good results for the estimation of forest dominant height on different tests sites, errors ranging from 1.3 to 3.2m (5% to 21%) depending on test sites and forest species.

Local maps are useful for forest managers and to study forest stands. However, it is difficult to make the link between these high resolution maps and statistics on larger areas such as those provided by the national forest inventory. In this study we aim to apply the methodology on a national scale in metropolitan France, using GEDI data to replace field height measurements, since there are not enough field height measurements over the whole country on a single year and a single protocol to be used to calibrate a national model. The objectives are i) to provide useful maps that can be used at multiple scales by foresters, decision-makers and scientists, and ii) to consolidate a simple and adaptable processing chain for the estimation and mapping of forest structure parameters.

## 2. Matérials and methods

### 2.1. Study site

The study area is metropolitan France, whose forests cover 31% of the metropolitan surface. According to (FAO, 2020) and the French institute of geographic and forest information (IGN), the surface area of French forests increased by 20% between 1990 and 2020 (including 30% which are new plantations), and the growing stock volume increased by 50% in the same period. However, since 2012 tree growth has slowed while mortality and harvesting have increased by 54% and 20% (*La forêt en France : portrait-robot - Portail IGN - IGN*, 2021), hence the need for increased monitoring of forests.

According to the IGN statistics, French forest consists mainly of deciduous trees, covering 67% of forest areas. The French forest contains 190 tree species, nearly three quarter of all the tree species in Europe. The forest is divided almost equally between pure stands (where one tree species occupies more than 75% of the canopy in the dominant storey) and mixed stands. In mixed stands we can find for example 5 different tree species per 2,000 m² plot. The forests of northeastern France and the Massif Central (center) are the most diverse. The Landes massif (south-west), on the other hand, is a large massif of mono-specific stands, with maritime pine as the main species. The seven main tree species are oak (with 44% of hardwood volume), beech and chestnut for hardwoods, and maritime pine, Scots pine and spruce for conifers. This diversity can be explained by the variety of environments. The National Forest Inventory (NFI) differentiates 86 sylvo-ecoregions in metropolitan France (Cavaignac, 2009), within which forests and/or environmental factors are close. The division into sylvo-ecoregions takes into account biogeographical factors (climatic, geological, topographical, pedological, floristic and landscape variability) that determine forest production and the distribution of major forest habitat types. The 86 sylvo-ecoregions are grouped into 11 GRECO regions (Grande Région ECOlogique, i.e. large ecological region). Sylvo-ecoregions and GRECO regions are represented in Figure 1, sylvo-ecoregions are labelled with a letter (corresponding to the GRECO region) and two numbers.

Today, private forests account for three quarters of forest area. This proportion has grown steadily in recent decades since the growth in forest surfaces results from land abandoned by agriculture and pastoralism. Overall, in western France, the proportion of private forest is significantly higher than the national average. The Grand Est region is the only one where private forest is in the minority (44%). Figure 1 shows the public (red) and private (blue) forest areas in France.

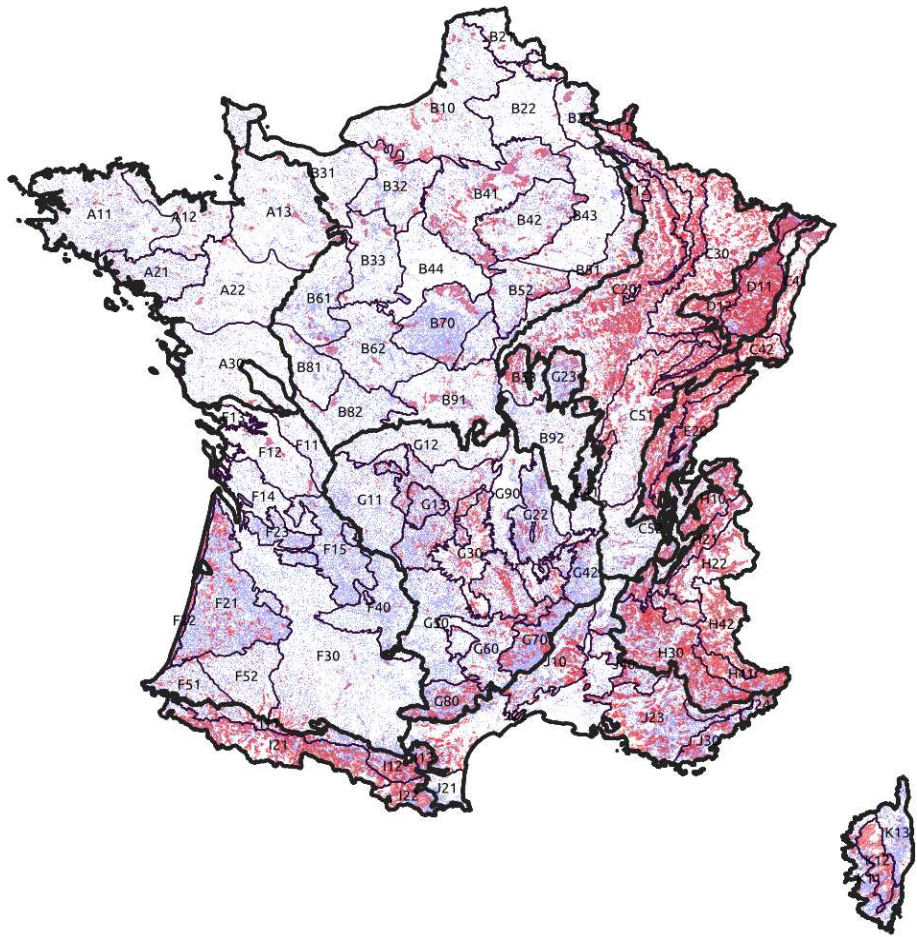

*Figure 1. Forest areas in France. Red: public forests, blue: private forests; GRECO regions are delimited in black with a thick line, sylvo-ecoregions are delimited in black and labelled with a letter and two numbers.*

## 2.2. Reference data

### 2.2.1. Training data

GEDI data are used in this study as reference data for the calibration of our height estimation method. GEDI is a laser instrument on board the International Space Station. It is equipped with three lasers, two of them are at full power beams and the other is split into two beams (called coverage beams), producing a total of four beams and eight ground tracks (see illustration in Figure 2.a). The footprints have an average diameter of 25 m and are separated by 60 m along the track and 600 m cross track. We have downloaded the L2A level data from November-December 2019 to November 2020 which provides, among other metrics, the relative height measurements. To assess its relevance for measuring tree height, we compared various relative height metrics and GEDI parameters with CHM maps supplied by the National Forest Office (ONF) for diverse sites (Morin *et al.*, 2022). We chose to use only full-power beams (not coverage beams) and good quality signal to noise ratio (sensitivity threshold > 0.95) to have good quality reference points. Across the whole of France, we end up with around 1 300 000 broadleaved and 500 000 coniferous GEDI samples. We chose the RH98 metric (relative height measurement at 98% energy) which showed good agreement on the CHM maps comparison (the MAE ranges from 1.5 to 3m depending on the test sites). Figure 2.b shows an example of the GEDI footprints in the Orleans forest.

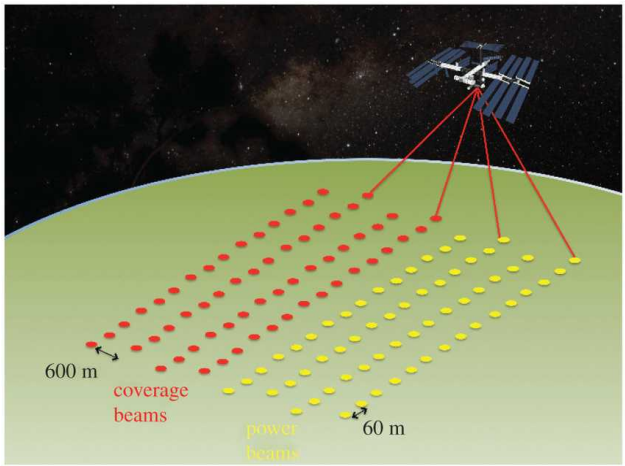 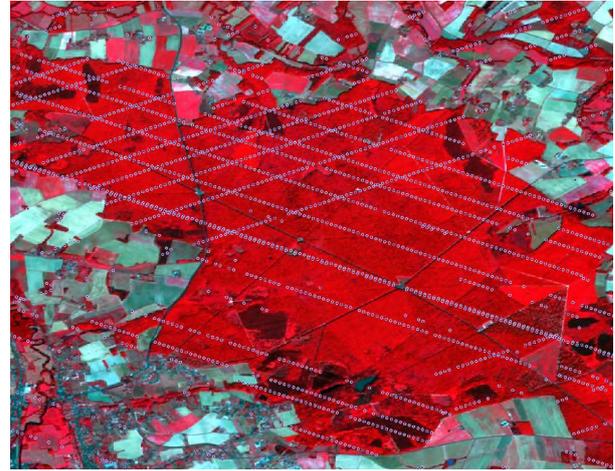

a) Illustration of GEDI sampling on the ground (source: Royal Society Open Science)    b) Example of GEDI tracks in the Orléans forest

*Figure 2. a) Illustration of GEDI sampling. b) false colour image (active vegetation in red), blue plots are GEDI samples.*

### 2.2.2. Validation data

The validation dataset includes 1) height measurements over forest plots carried out by French NFI (IGN), 2) forest measurements with volume and above-ground biomass estimations carried by the national forest office (ONF) on public forests in 2020, and 3) canopy height models (CHM) derived from airborne Lidar (ALS) measurements over five diversified sites. These data were used only for the independent validation of the height maps derived from GEDI reference data and the Sentinel-1/Sentinel-2/ALOS-2 PALSAR2 satellite images. Figure 3 shows the distribution of the NFI plots and CHM dataset used in this paper.

1) NFI plot measurements: the NFI covers the whole mainland France with a systematic sampling on a 1 km square grid. A subsample of these points are visited in the field each year. On the forest plots, different information is gathered on concentric plots with radius from 6 to 20m depending on the data of interest. This measurement dataset is representative of the French forest. Among other measurements, height is measured for a sample of trees representative of diameter categories within the plots. The dominant height is defined as the average height of the 100 largest trees per hectare, the calculation is based on the categories of trees measured inside the 20m plot. The volume is calculated on the basis of tree height and DBH and shape coefficient depending on tree species. The above-ground biomass (AGB) was derived from the NFI volume according to the ratios given in the CARBOFOR project (Loustau *et al.*, 2004; Loustau, 2010): 0.89 tons/m³ for broadleaved forests and 0.59 tons/m³ for coniferous forests.

2) ONF plot measurements: the ONF carried out measurements in 2020 in public forests on several study sites presented in blue in Figure 3. The height, diameter and number of trees per hectare were measured. The volume was estimated using the equations by species given in the EMERGE project (*Rendez-vous techniques de l'ONF - n° 44*, 2014), and the biomass was calculated using infradensity by species from (Dupouey, 2002). 745 broadleaved plots and 486 coniferous plots are available over 6 study sites.

3) CHMs: the airborne lidar measurements were carried out by the ONF in 2020 in 5 diversified sites : Andaine, Carcans-Hourtins, Deodatie, Mouterhouse and Lajoux-Fresse. The Andaine site is on a flat area in north-east France with heavily managed stands, the Carcans-Hourtins site in south-west France (Landes de Gascogne) is mainly composed of maritime pine stands. The other sites are in mountainous areas in the northeast of France (Jura and Vosges). Canopy height maps (CHM) at 1m spatial resolution were derived from ALS data. We resampled these CHM on a 10m×10m grid using the maximum height to compare to our height map. ALS characteristics are presented in table 1.

*Table 1. ONF Airborne Lidar characteristics for the 5 sites.*

|  | Scanning density (points/m²) | Vertical accuracy (cm) | Absolute positional accuracy (cm) |
|---|---|---|---|
| Andaine | 22 | 8 | <15 |
| Carcans-Hourtins | 31 | 10 | <15 |
| Deodatie | 8 | 5 | <10 |
| Mouterhouse | 17 | 5 | <10 |
| Lajoux-Fresse | 31 | 4 | <20 |

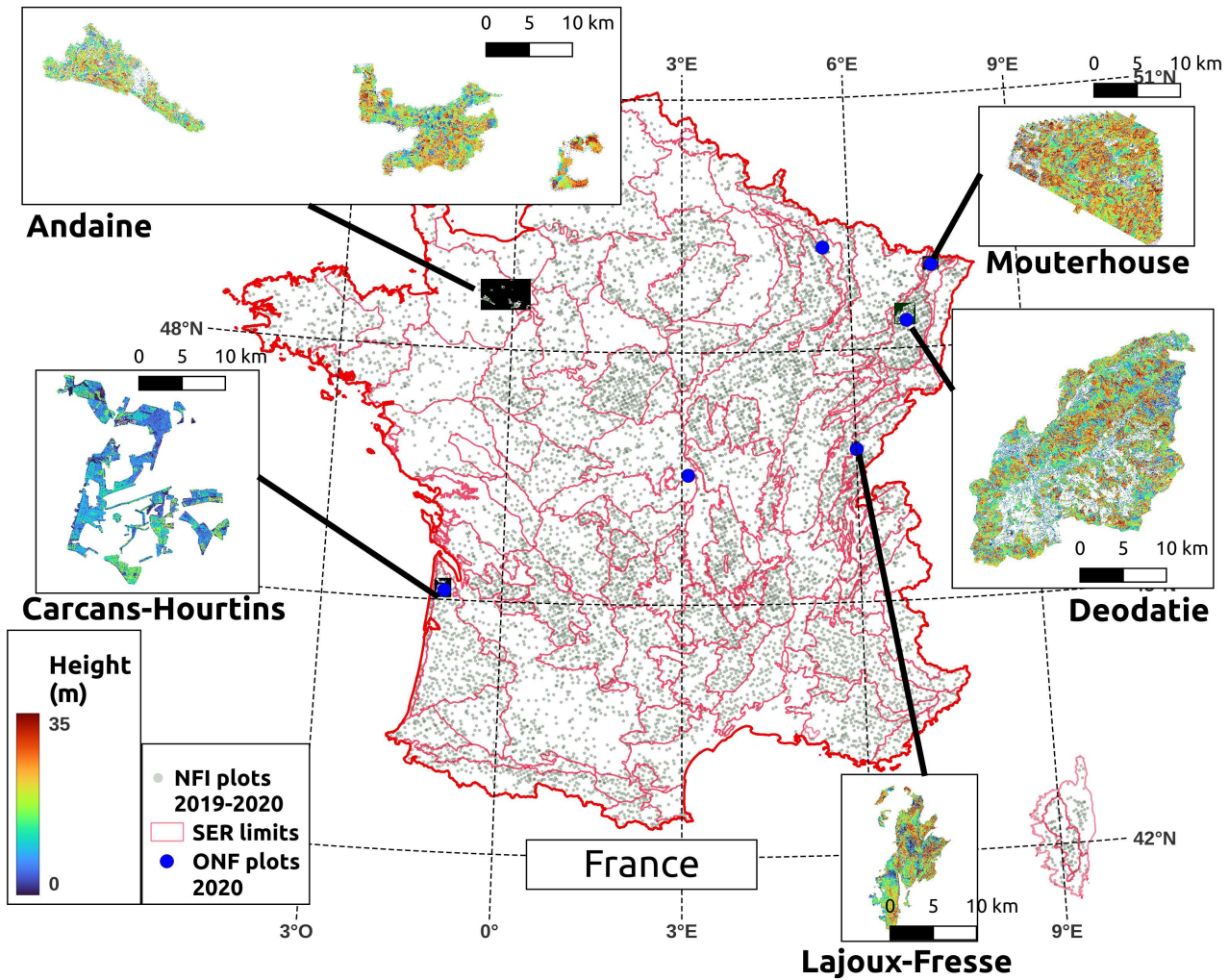

*Figure 3. Validation data: ALS-CHM from ONF on 5 test sites, NFI plots (grey points), ONF plots (blue points) and SER (boundaries in red).*

## 2.4. Satellite imagery

This section provides information on the extraction of image features used for forest height prediction. These features come from Sentinel-1 and Sentinel-2 satellites (European Copernicus program), and ALOS-2 PALSAR-2 (JAXA, Japanese Aerospace Exploration Agency).

The Sentinel-1 satellites provide C-band synthetic aperture radar (SAR) images acquired every 6 to 12 days at 10m pixel size. We pre-processed GRD (ground range detected) images for calibration, orthorectification, radiometric correction to tackle negative effects of local slope and resampling to superimpose on Sentinel-2 images, using the S1Tiling library[1]. We exploited data acquired in 2020 during the winter period (December, January, February) and the summer period (June, July and August), where the phenology is the most stable and spatially homogeneous (Morin *et al.*, 2019). We processed ascending (~6pm acquisition time) and descending (~6am acquisition time) orbits separately, obtaining winter/summer ascending/descending average values. Ascending and descending averages were also combined to attenuate the differences in climatic conditions between morning and evening (presence of dew, wind) that affects the radar signatures. At the end of the Sentinel-1 data processing, we have 4 backscatter features (winter and summer VH and VV polarisations).

The Sentinel-2 satellites provide optical images acquired every 5 to 10 days under the same acquisition conditions. The reflectances measured are in the visible, near and mid infrared. The level 2A images were downloaded from the Theia platform[2]. Images are ortho-rectified and corrected for atmospheric effects with automatic cloud detection. We processed data acquired at the same Sentinel-1 winter and summer periods, temporal composites were computed using the WASP software[3]. Cloud-free pixels were averaged using weighting factors. We used the Sentinel 2 spectral bands listed in Table 2, from which were derived four spectral indices: brightness (BI), vegetation activity (NDVI), wetness (NDWI) and chlorophyll (ND56). At the end of the Sentinel-2 data processing, we have 8 spectral features (winter and summer BI, NDVI, NDWI and ND56 indices).

*Table 2. Sentinel-2 spectral bands information and spectral indices extracted.*

| Bands | Central wavelength (μm) | Resolution (m) | Bandwidth (nm) |
|---|---|---|---|
| Band 2 – Blue | 0.490 | 10 | 65 |
| Band 3 – Green | 0.560 | 10 | 35 |
| Band 4 – Red | 0.665 | 10 | 30 |
| Band 5 – Vegetation red edge | 0.705 | 20 | 15 |
| Band 6 – Vegetation red edge | 0.740 | 20 | 15 |
| Band 7 – Vegetation red edge | 0.783 | 20 | 20 |
| Band 8 – NIR | 0.842 | 10 | 115 |
| Band 8A – Narrow NIR | 0.865 | 20 | 20 |
| Band 11 – SWIR | 1.610 | 20 | 90 |
| Band 12 – SWIR | 2.190 | 20 | 180 |

$$BI = \sqrt{band\,2^2 + band\,3^2 + band\,4^2}$$

$$NDVI = \frac{band\,8 - band\,4}{band\,8 + band\,4}$$

$$NDWI = \frac{band\,8A - band\,11}{band\,8A + band\,11}$$

$$ND56 = \frac{band\,6 - band\,5}{band\,6 + band\,5}$$

The L-band ALOS-2 PALSAR-2 annual mosaics at 25m pixel size and HV/HH polarisation, produced by JAXA, were used in this study. We downloaded the 2015-2021 images and applied multi-image speckle noise filtering (Bruniquel and Lopes, 1997; Quegan and Yu, 2001), reprojection and retiling to superimpose at the same pixel grid of Sentinel-1 and 2 images. At the end of the ALOS-2 PALSAR-2 data processing, we have 2 backscatter features (annual HV and HH polarisations).

Spatial texture indices have been generated from both Sentinel-2 10m spatial resolution spectral indices (NDVI, BI) and Sentinel-1 polarisations (VH, VV), using a Grey Level Co-occurrence Matrix (GLCM) with the Orfeo Toolbox library (Grizonnet *et al.*, 2017). Three second order Haralick metrics were calculated on GLCM: homogeneity (also called Inverse Difference Moment), contrast (also called Inertia), and Haralick correlation. Based on a previous sensitivity analysis (Morin, 2020), the following parameters were used to obtain the textural metrics for further processing: offset set to 1 pixel; 3 pixels radius; 100 grey levels (number of bins); minimum and maximum calculated separately between Sentinel-2 indices and dates, and between Sentinel-1 polarisations; and orientations 0°, 45°, 90° and

---

[1] https://s1-tiling.pages.orfeo-toolbox.org/s1tiling/latest/
[2] https://theia.cnes.fr
[3] https://github.com/CNES/WASP

135 ◦ were computed then averaged. In total, we have 12 texture features for Sentinel-1 and 12 for Sentinel-2.

## 2.5. Ancillary products

Forest stratification, for example by geographical areas or by species, was tested and used to stratify the data used as inputs of our height estimation model. The use of forest stratification improved the height estimations (Morin, 2020, 2022) because the height of a tree is influenced by several factors: species, type of soil, climate, topography, and forest silvicultural practices.

For the spatial/geographical stratification, we used the IGN's sylvo-ecoregions (Cavaignac, 2009) freely available (see 2.1) to learn a model for each zone. For stratification by species, we used the Dominant Leaf Type (DLT) map of the Copernicus HRlayers from 2018 [4] which classifies deciduous and coniferous trees. DLT map is also freely available and produced throughout Europe.

## 2.6. Estimation method

Our processing chain was developed previously in (Morin *et al.*, 2019, 2022; Morin, 2020) with successful results at local scale using satellite images and ground measurement or GEDI samples as reference samples. In this study we have slightly adapted the method in order to improve the estimates (modification of the regression algorithm) and to allow scaling up (learning by SER, and with no need for feature selection). Figure 4 presents the processing chain workflow. The steps are:

1) First, the calibration dataset consists of satellite image features and GEDI samples. For each GEDI footprint the corresponding values in feature images are extracted and averaged (with 4 to 6 pixels per footprint) in order to build a general learning table. The dataset is then stratified: the table is separated into several learning tables according to sample location in the SER and the type of forest species (broadleaved or coniferous).

2) Model learning is done using an algorithm that combines Random Forest (RF) and linear regression, called Linear Forest Regression (LFR) and available online[5]. LFR algorithm allows to keep the computational efficiency of Random Forest while improving the performances, especially for low and high forest heights. After preliminary tests, an optimisation is done for each learning tables on 4 key parameters: n_estimators [50,100,200], max_features ["auto","sqrt"], max_depth [10,20,30] and min_samples_split [15,25,35]. The final models are obtained in two steps. i) First, for each dataset (SER/forest) a K-fold cross-validation is done using 10 folds. Based on the result, outliers are removed from the learning tables. Outliers are defined as follows: more than 10 meters error, or more than 100% relative error if reference height is under 10 meters. This step allows to take away samples with large geolocation errors (and close to non-forest land cover or forest changes) or height estimation error, especially for the GEDI samples where only 1 peak is detected. ii) Then, the new learning tables are fully used for learning the models that will be used to produce the maps.

3) Prediction maps are obtained by applying models on the image feature stack, in correspondence with the forest class (SER and species) of each pixel.

4) The conversion of height into volume and AGB was achieved using a power law $Y=a \times Height^b$ between height and volume or AGB. All NFI plot measurements since 2005 are available online[6] without exact coordinates. Power laws for volume and AGB were fitted on deciduous and conifer classes between dominant height (Hdom) and volume or AGB of the NFI dataset. The parameters to be optimised are bounded in order to prevent saturation of the relationship. 82417 plots were available for broadleaved forests, and 27588 for coniferous forests. Once the power law parameters were adjusted, we applied them to our height map. Tests were carried out to optimize power laws by SER, without improvement (see results section).

---
4   https://land.copernicus.eu/pan-european/high-resolution-layers/forests/dominant-leaf-type/status-maps/dominant-leaf-type-2018
5   https://github.com/cerlymarco/linear-tree ; uses the scikit-learn implementation
6   https://inventaire-forestier.ign.fr/dataIFN/

5) Finally, an independent validation for the height on a local scale is done using the CHM reference maps. Height values are compared pixel by pixel and we computed validation statistics. In addition, we extracted the values from the prediction maps on the 2020 ONF plots and compared them to the measured values. Finally, at national scale, NFI plots were crossed with our height map thanks to the IGN so we have NFI height, volume and AGB compared to our map values per NFI plot for the years 2019 and 2020. The same statistics than for CHM were computed. Validation statistics are: coefficient of determination r² (Pearson), the root mean squared error (RMSE), the relative RMSE (rRMSE in %, RMSE divided by the mean of the reference values), the mean absolute error (MAE), the relative MAE (rMAE in %, MAE divided by the mean of the reference values), and the mean bias (sum of all errors).

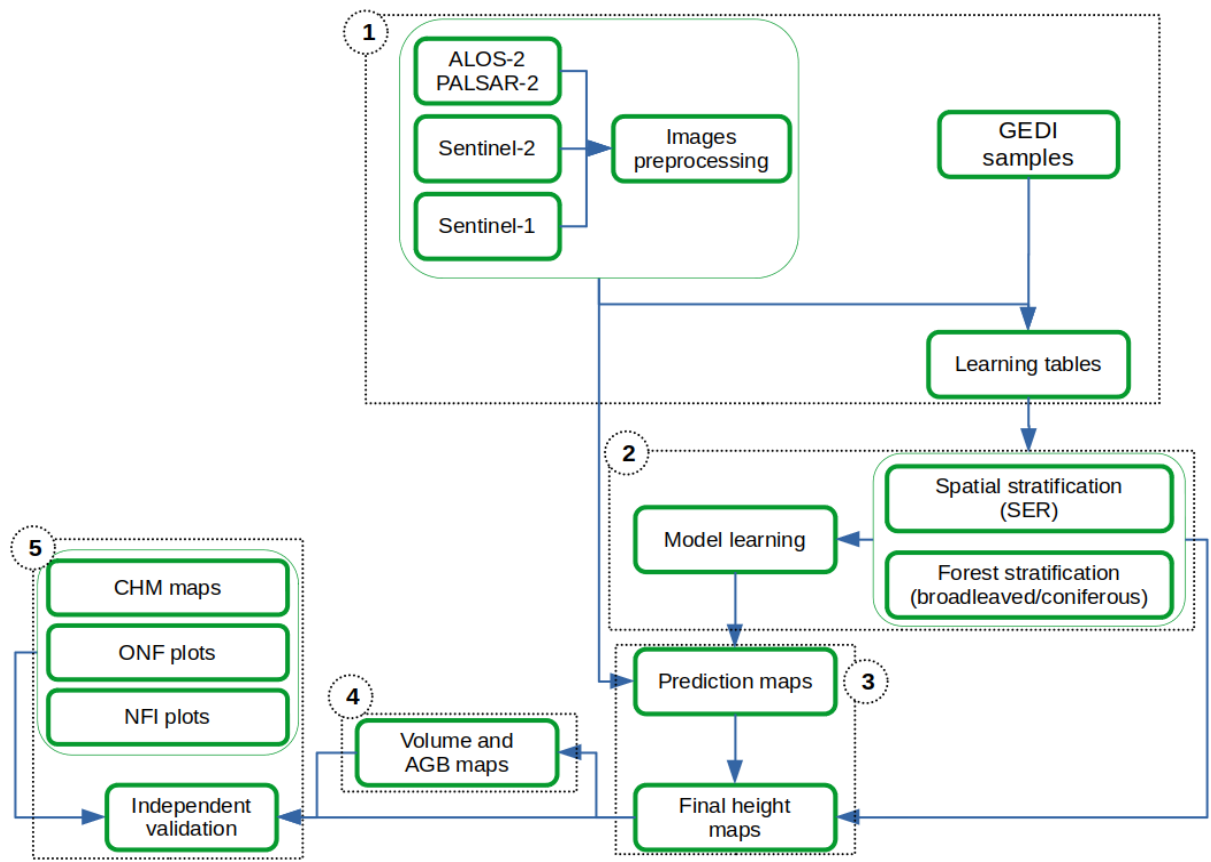

*Figure 4. Flowchart of the height estimation method developed in this study.*

## 3. Results

### 3.1. Heigh map

The methodology presented above allowed us to produce a map of forest canopy height at 10 m spatial resolution on mainland France in 2020. Figure 5 presents this map on a national scale. We can see different patterns according to the territories: low and quite open forests in the Mediterranean zone (south-east), regular patches with predominance of low to medium heights in the Landes de Gascogne (close-up in the south-west) typical from forest plantations (of maritime pine in this case), large and irregular forest areas with predominance of high heights in the mountainous zones (close-up in the north-east), and various configurations (regular and irregular patches, low and high heights) in the plains (close-up of center north). Local close-ups in Figure 5 show that forest stands with different management and height are easily differentiated.

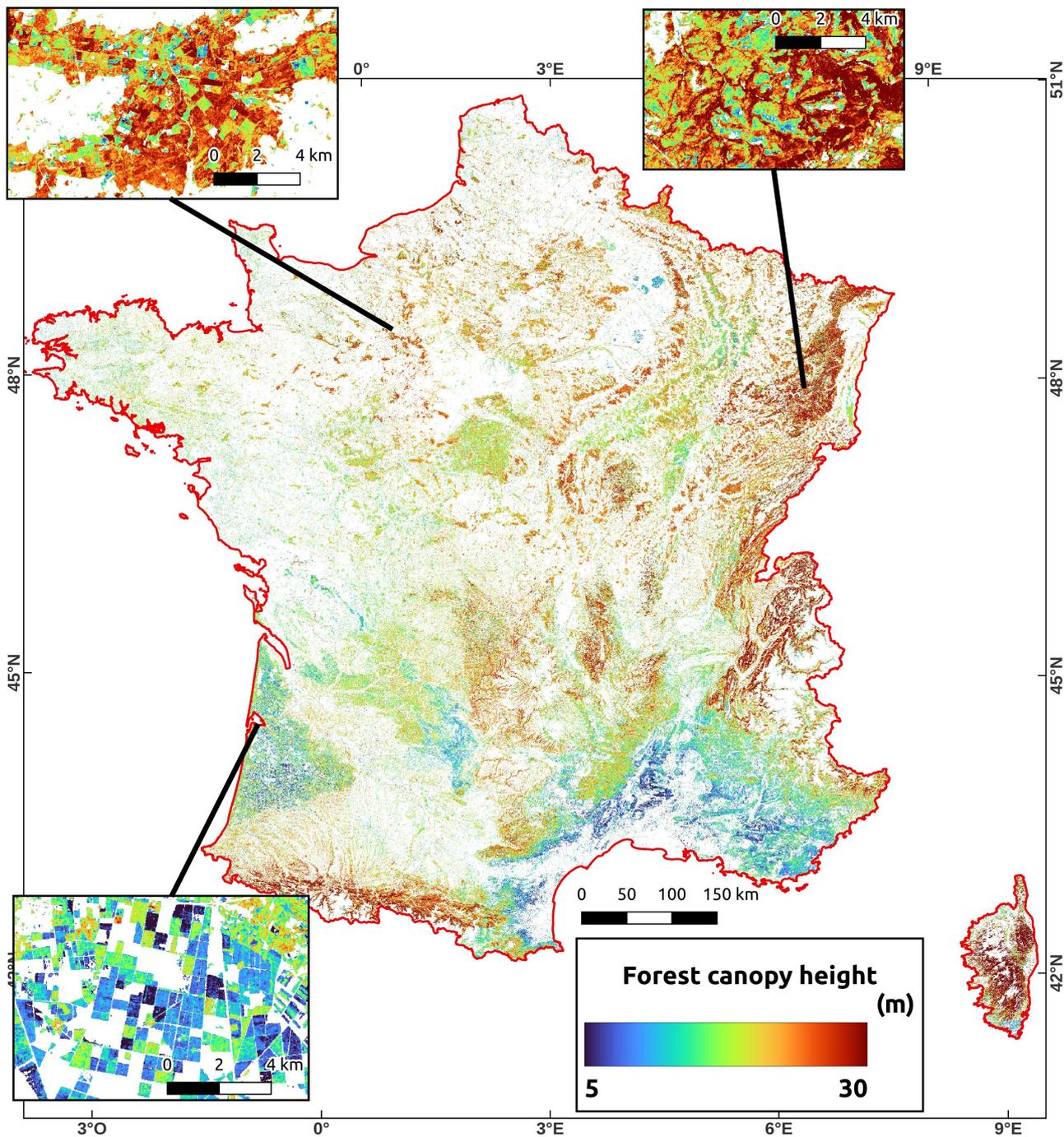

*Figure 5. France canopy height map at 10m spatial resolution, produced using the method developed in this study.*

### 3.1.1. Height validation with NFI plots

Height estimations maps were crossed with 2019 and 2020 NFI plots by the LIF (IGN). Figure 6 shows the validation statistics. The mean absolute error (MAE) is around 3.7m for both broadleaved and coniferous forests. Coniferous height estimates have a better correlation with the reference ($r^2$=0.6 against 0.45 for broadleaved), partly because they have a wider distribution of height values, also the saturation of high values seems less strong.

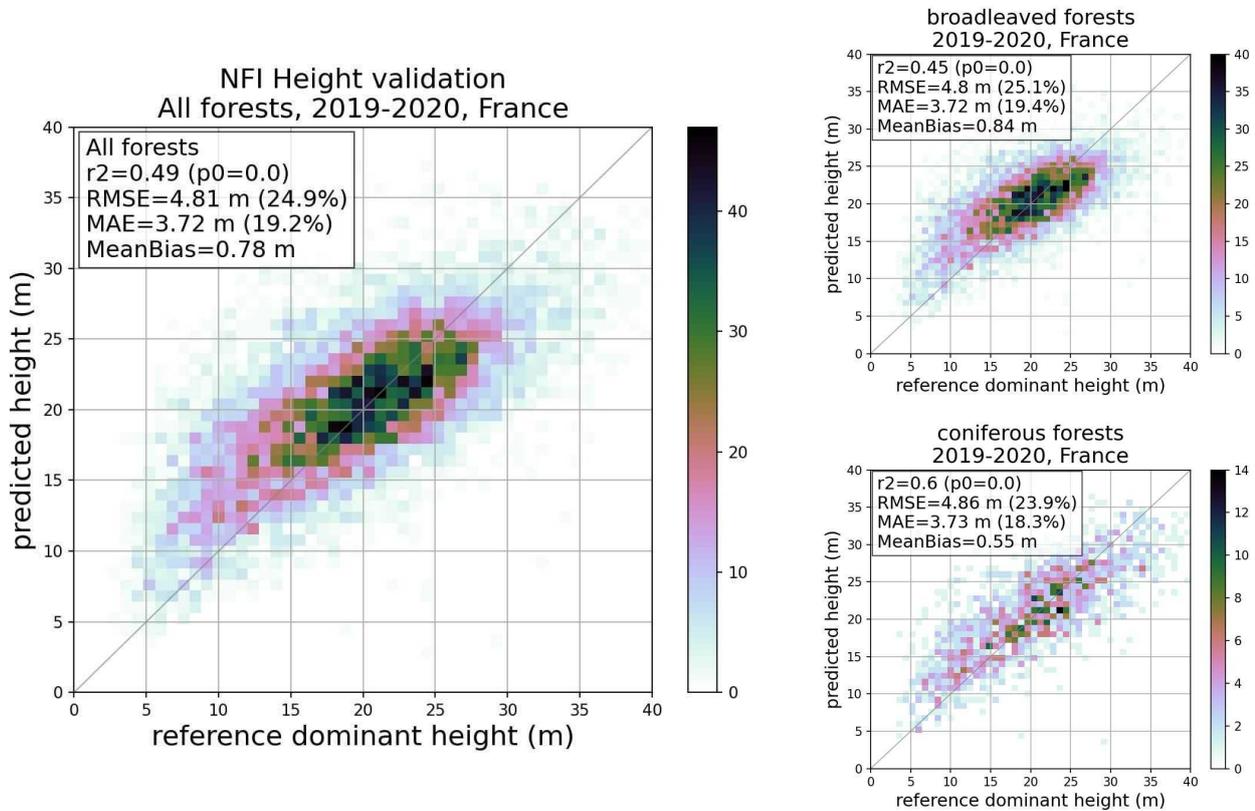

*Figure 6. Validation of height estimations on NFI plots throughout France.*

### 3.1.2. Height validation with ONF plots

The height map was also crossed with 2020 ONF plots over 6 study sites in France. The distribution of measured height values is a little different from the NFI dataset over the whole of France. There are few low heights, and more heights above 25-30m. Figure 7 shows the validation statistics. The MAE is 4.93 m (23.8%) for broadleaved and 3.32 m (14.1%) for coniferous. Again, the predictions for coniferous seem better than for broadleaved forests. The heights of broadleaved forests between 12 and 20 m are overestimated.

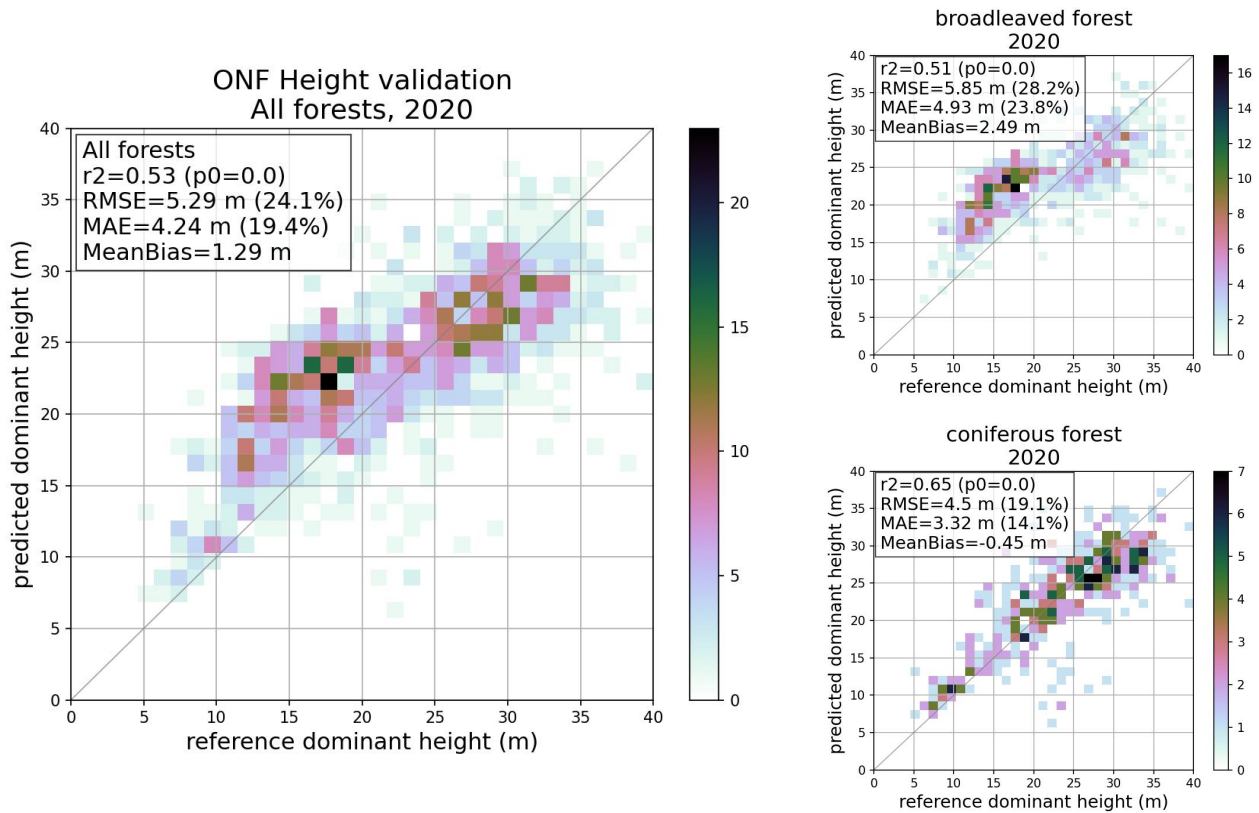

*Figure 7. Validation of height estimations on ONF plots throughout 2020 measurement sites.*

### 3.1.3. Height validation with CHMs

We compared the predicted height map with ALS CHM on 5 sites. The figure 8 shows visual comparison on two small areas: plains on the top line, and mountains on the bottom line. We can see that the patterns are well represented, even if we observe some local errors (e.g. center of the bottom line images). The 10m spatial resolution is well conserved in our product, despite some degradation expected from spatial texture indices.

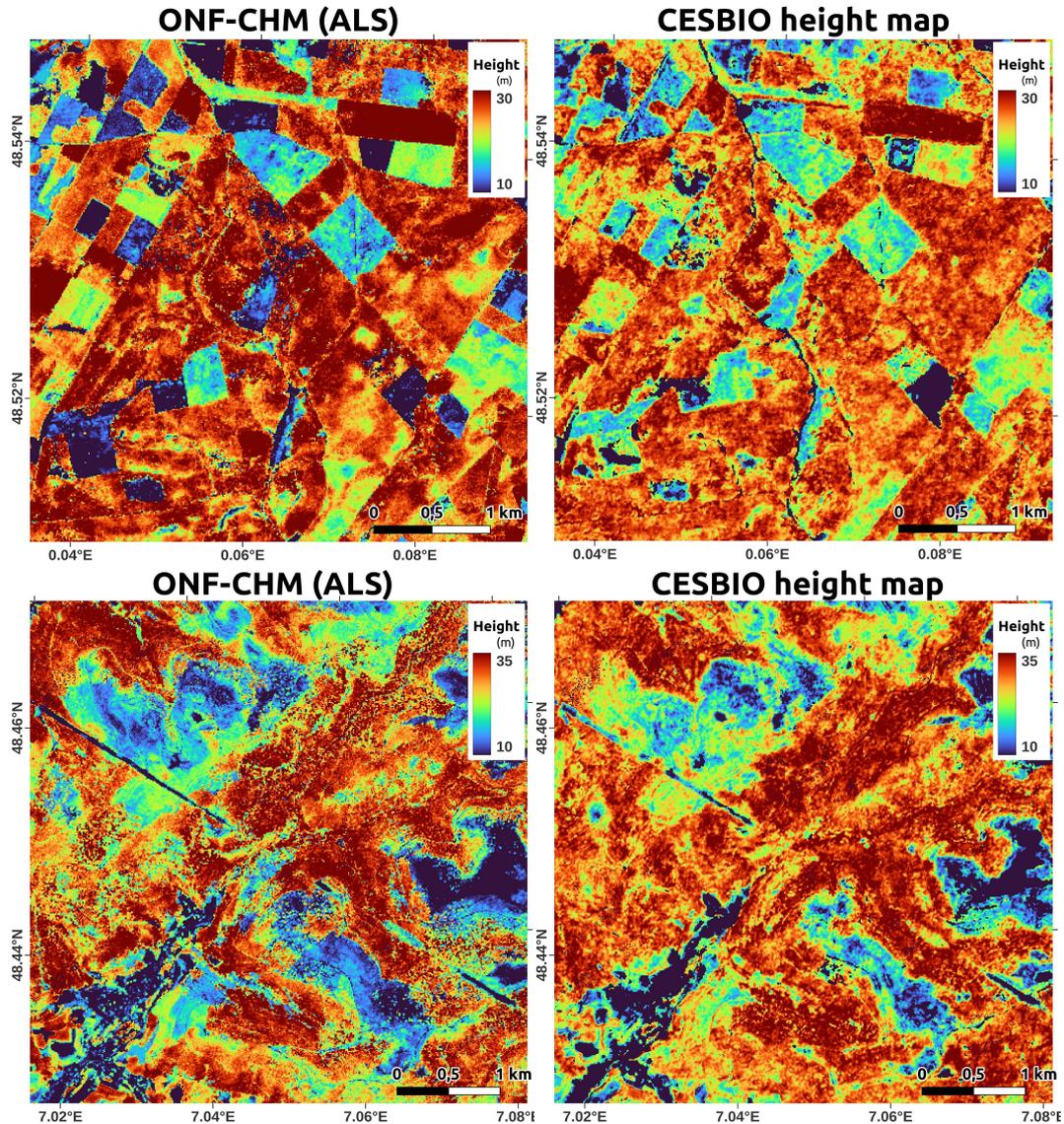

*Figure 8. Maps comparison between CHMs (left column) and the predicted height map (right columns) on a study site in the plain (top line) and another in the mountains (bottom line).*

The figure 9 shows the comparison statistics between our height map and the CHMs. Heights are compared pixel to pixel. Although not representative of all French forests, these local Lidar maps show different types of forests, sylvicultural and environmental conditions. The results are very close to the previous validation with NFI plots (figure 6). The relative MAE is around 20%. Height prediction works slightly better in coniferous forests. Saturation is more pronounced for broadleaved. Table 3 shows validation statistics on all 5 test sites separately. Absolute errors range from 17 to 32% for deciduous forests, and 15 to 19% for coniferous forests. These results remain accurate on mountainous sites despite the slopes, which introduce noise and can interfere with the relationship between the satellite signal and forest height.

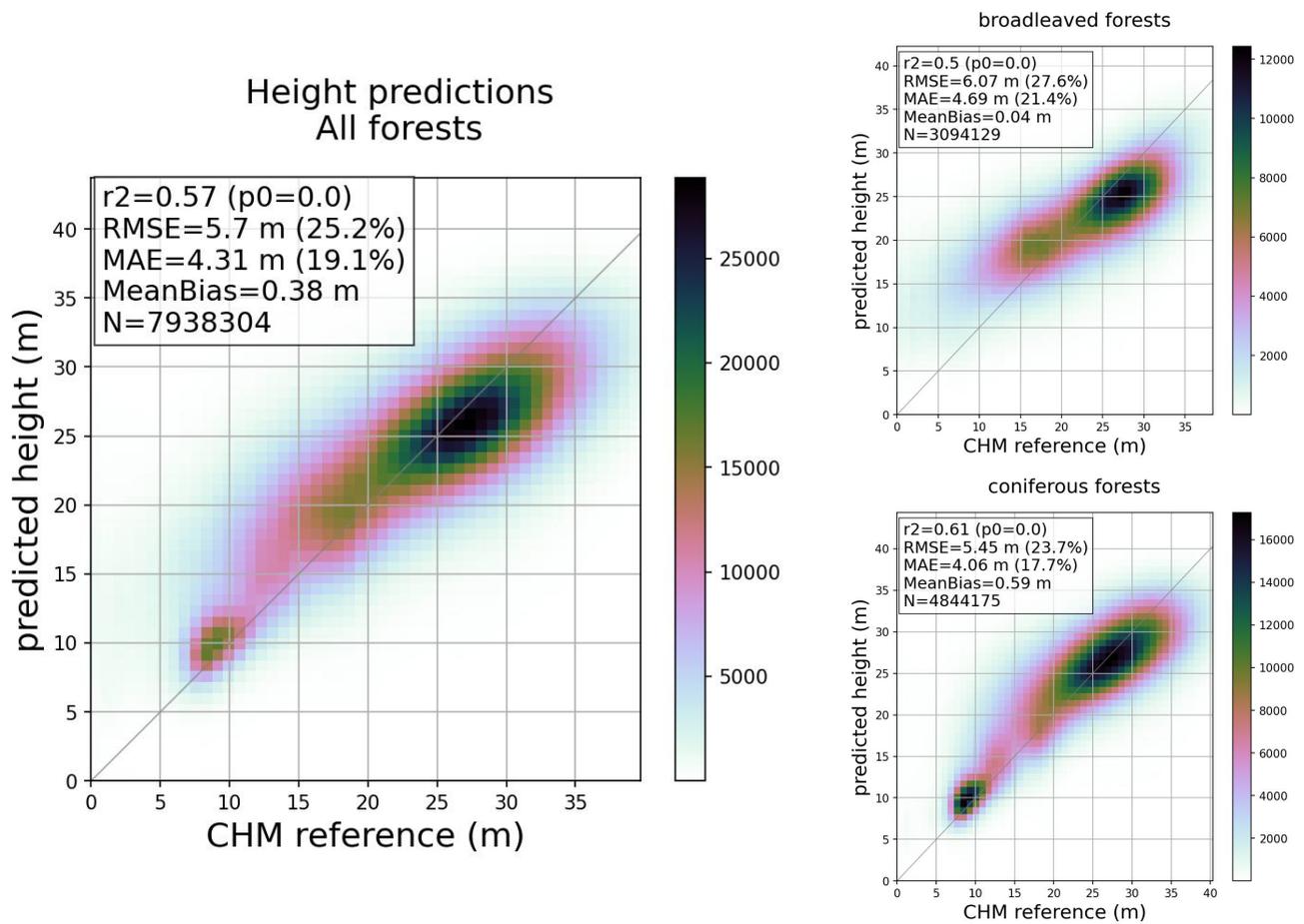

*Figure 9. Validation of height estimation: pixel-based comparison between predicted height map and CHM maps. All 5 sites are gathered.*

*Table 3. Validation statistics for the comparison of the height map with the CHMs on the 5 sites.*

| | | **Height map validation** | | |
|---|---|---|---|---|
| site | forest | r2 | MAE (m) | rMAE (%) |
| **Andaine** managed forest, plain | All forests | 0.48 | 4.1 | 18.0% |
| | Broadleaved (66%) | 0.49 | 4.1 | 19.0% |
| | Coniferous (34%) | 0.45 | 3.9 | 16.2% |
| **CarcansHourtins** coniferous plantations, plain | All forests | 0.54 | 2.3 | 19.0% |
| | Broadleaved (6%) | 0.28 | 3.4 | 25.1% |
| | Coniferous (94%) | 0.56 | 2.2 | 18.5% |
| **Deodatie** mountainous area | All forests | 0.48 | 5.0 | 21.2% |
| | Broadleaved (24%) | 0.32 | 5.7 | 32.4% |
| | Coniferous (76%) | 0.44 | 4.7 | 18.7% |
| **Mouterhouse** mountainous area | All forests | 0.40 | 4.4 | 17.1% |
| | Broadleaved (11%) | 0.44 | 4.6 | 17.5% |
| | Coniferous (89%) | 0.33 | 4.1 | 16.3% |
| **LajouxFresse** mountainous area | All forests | 0.53 | 4.1 | 16.2% |
| | Broadleaved (65%) | 0.50 | 4.7 | 27.3% |
| | Coniferous (35%) | 0.43 | 4.1 | 15.3% |

## 3.2. Volume and biomass maps

Once the height maps are produced, we need to convert these forest heights into volume and biomass. The NFI data downloaded provides us with the height and volume for each plot measured since 2005, and AGB was derived from volume using ratios from (Loustau *et al.*, 2004). We used a power law to adjust the relationship between height and volume or biomass. Figure 10 shows the relationships and equations we calculated for broadleaved and coniferous. The correlations are strong, a little better for coniferous ($r^2=0.68$) than for broadleaved forests ($r^2=0.6$). Since AGB is derived directly from volume, the correlations are similar between volume and AGB.

We tested extracting different equations by SER or by GRECO, but the overall results were not significantly improved so we preferred to stay with a simple solution differentiated only between broadleaved and coniferous forests.

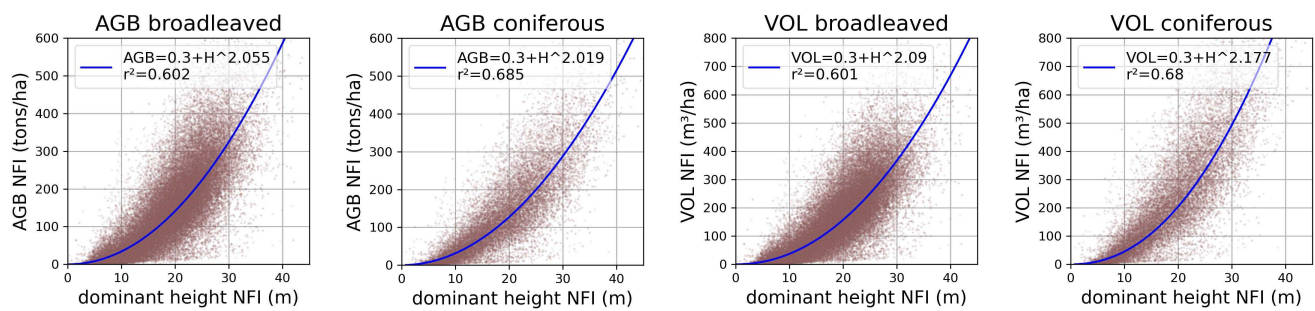

*Figure 10. Relationships between forest dominant height and above-ground biomass (AGB) and volume (VOL) within the French NFI dataset (2005-2021).*

After extracting the equations linking height and volume/AGB, we applied these equations to our height map in France. Figure 11 shows the map of forest above-ground biomass in France, at 10m spatial resolution for the year 2020. As for the height map in Figure 5, we can observe the distribution of forest biomass in France. The zooms in the figure 11 show different forests and we can see significant biomass dynamics in the stands.

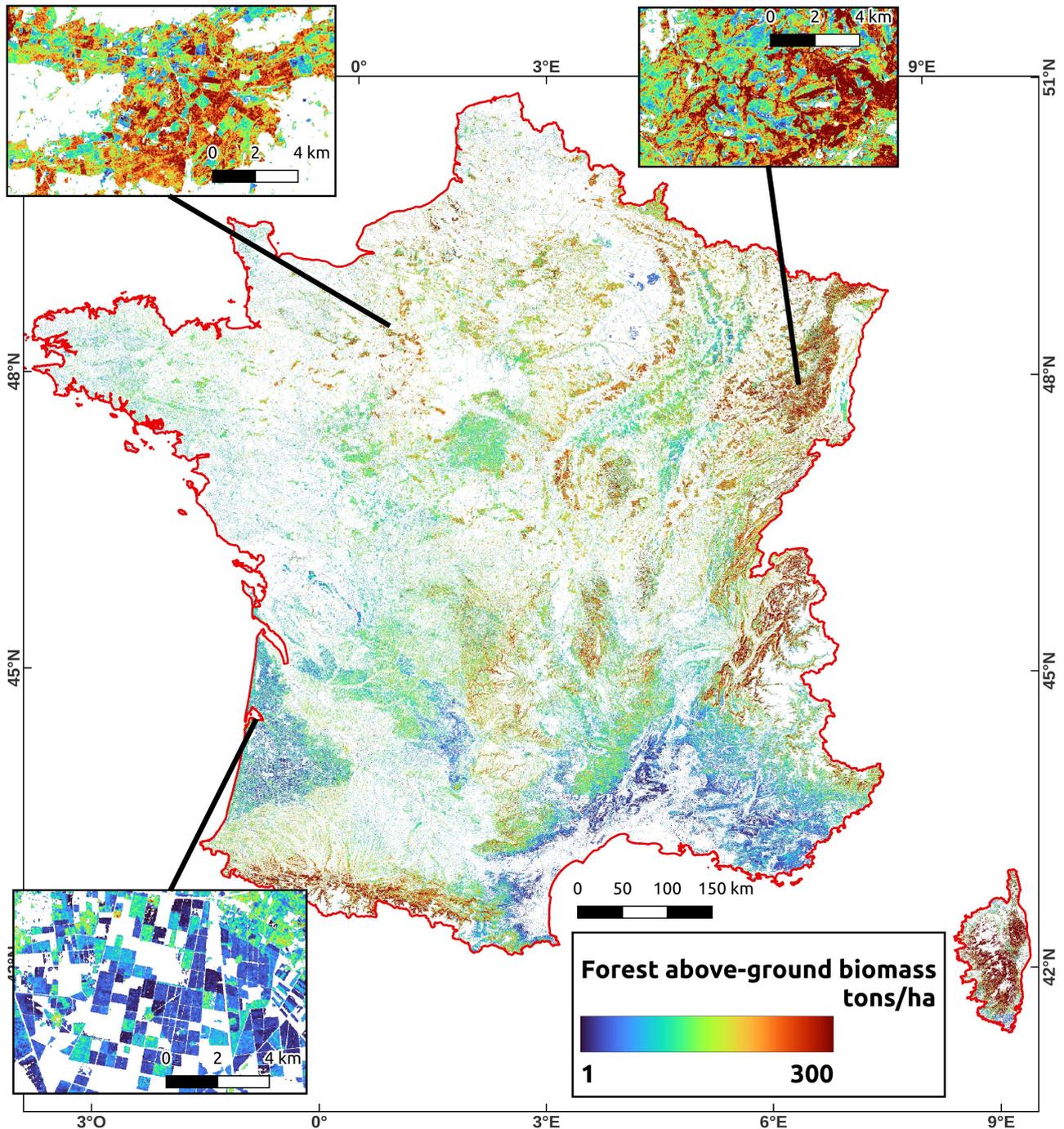

*Figure 11. France AGB map at 10m spatial resolution, derived from the height map produced using the method developed in this study.*

### 3.2.1. AGB and volume validation with NFI plots

In order to validate these volume and AGB maps, we also applied the allometric equations to the heights of our map which had been crossed with the 2019-2020 NFI plots. Figures 12 and 13 show the validation of AGB and volume estimations on 2019 and 2020 NFI plots. The precision is less good than for height validation, with an MAE of 45% (40% for coniferous). Indeed, the calculation of the AGB and the volume from the height introduces noise, and amplifies the saturation in particular for broadleaved. Nevertheless, these results remain interesting and allow us to observe the dynamics of biomass and volume.

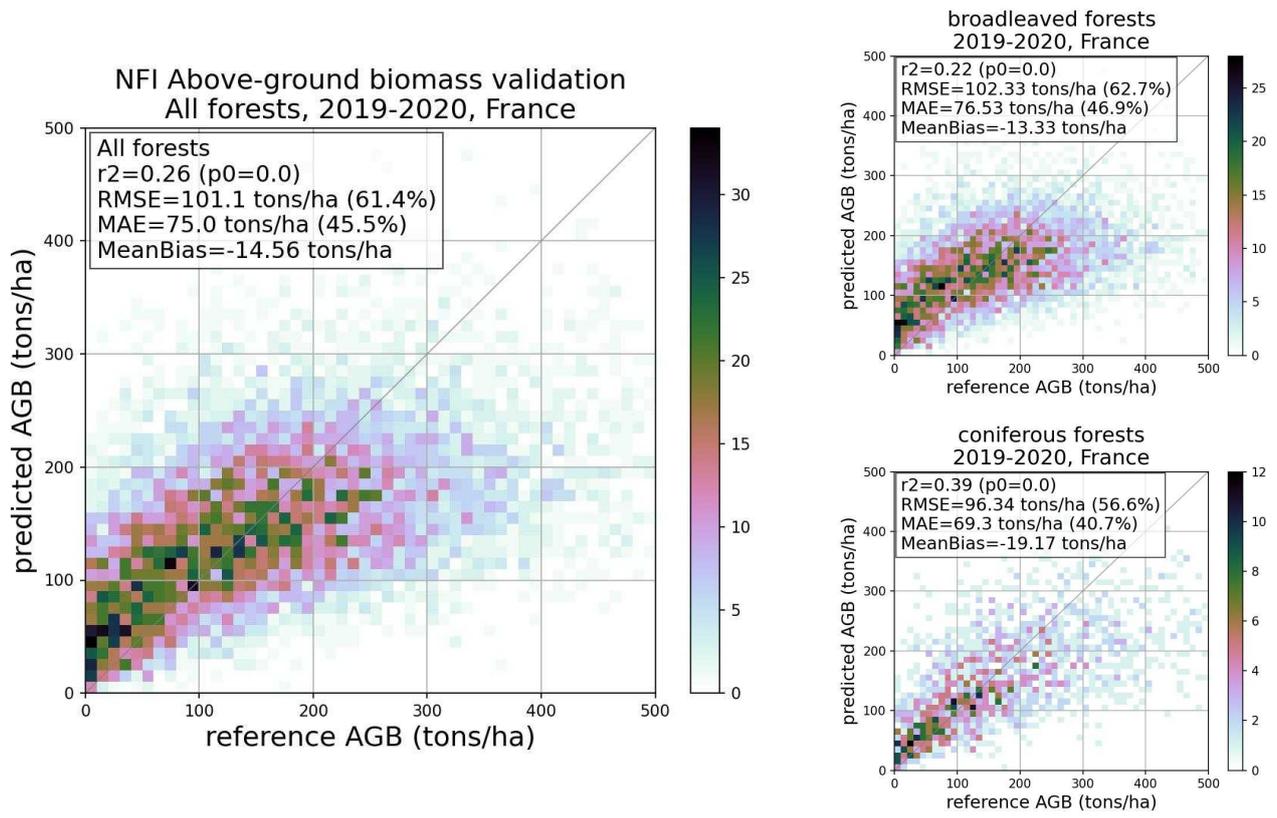

*Figure 12. Validation of above-ground biomass (AGB) estimations on NFI plots throughout France.*

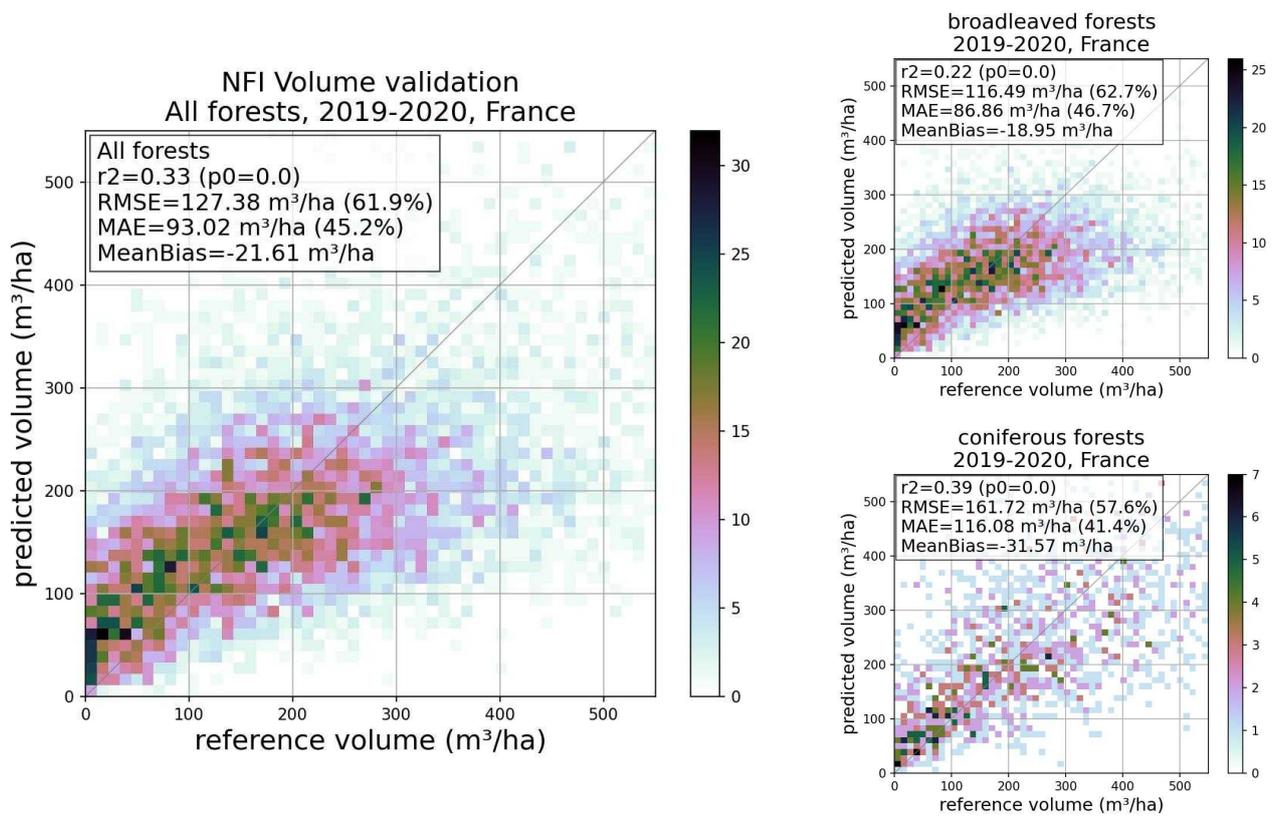

*Figure 13. Validation of wood volume estimations on NFI plots throughout France.*

### 3.2.2. AGB and volume validation with ONF plots

AGB and volume prediction maps were also crossed with ONF plots. Figure 14 and 15 shows the validation statistics for volume and AGB respectively. The relative MAE are around 40% for the AGB and 38% for volume estimations. The AGB seems slightly overestimated while volume is underestimated. The difference in AGB and volume calculation between this dataset and the national inventory may affect the results.

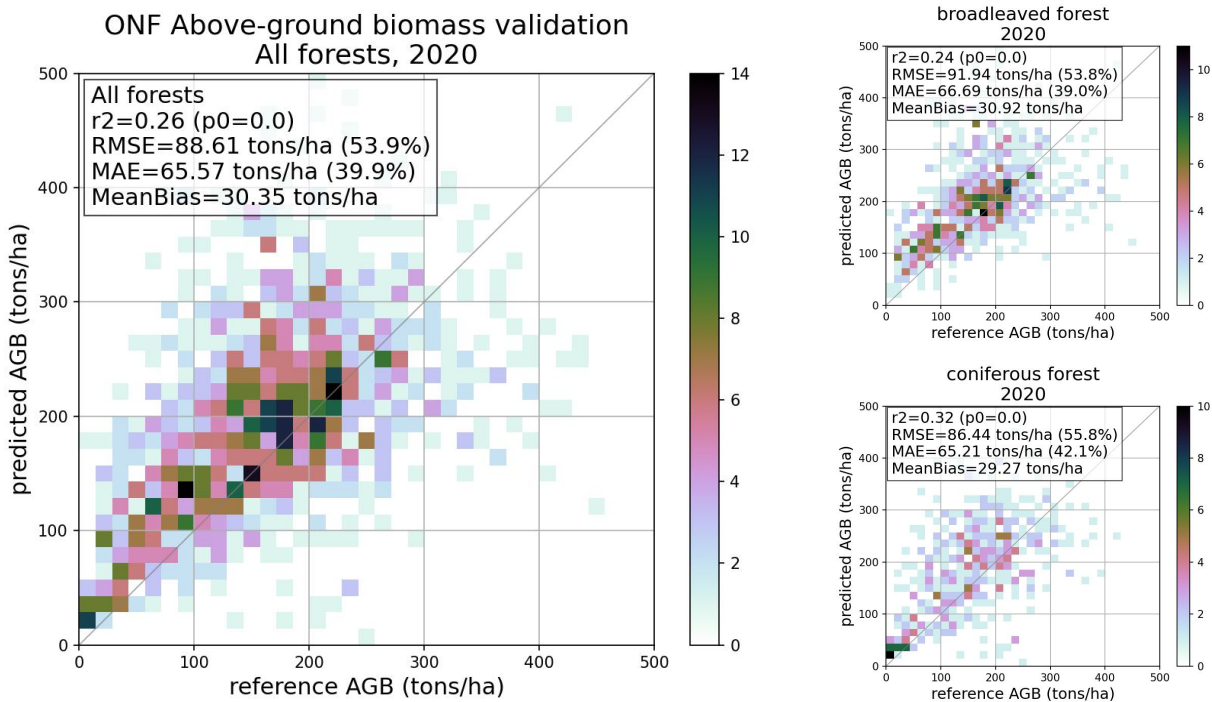

*Figure 14. Validation of above-ground biomass (AGB) estimations on ONF plots throughout 2020 measurement sites.*

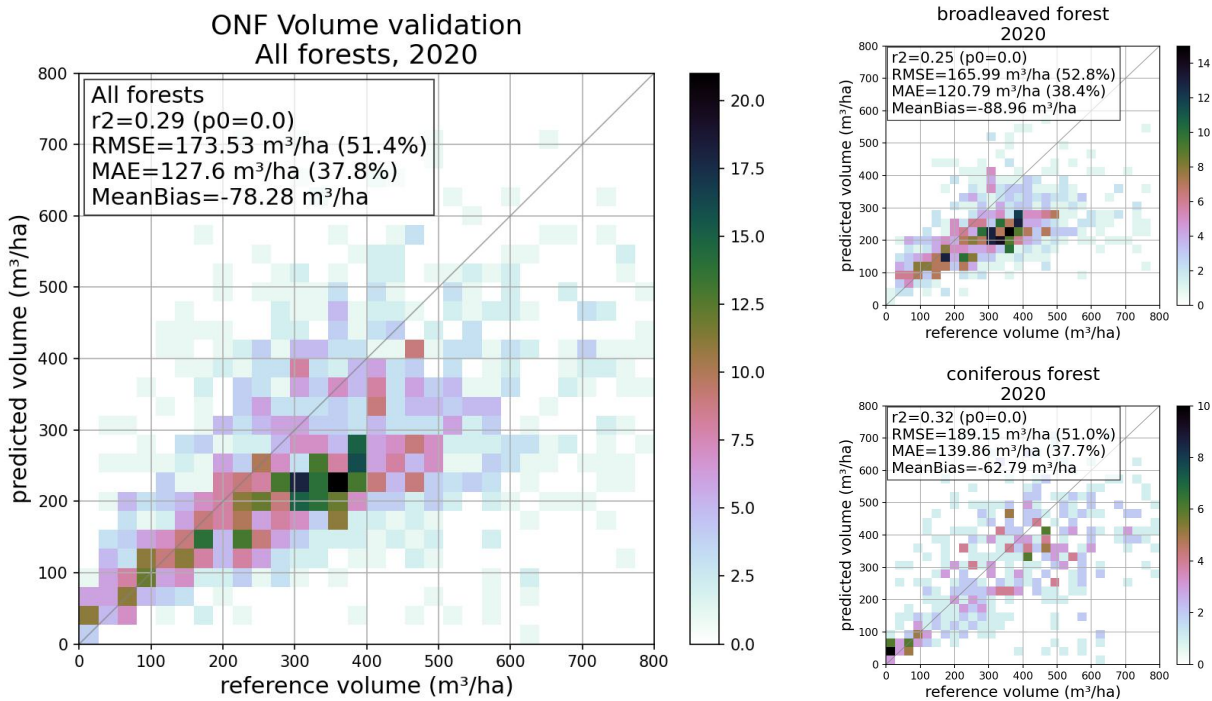

*Figure 15. Validation of wood volume estimations on ONF plots throughout 2020 measurement sites.*

# 4. Discussion and conclusion

The methodology presented in this document shows accurate overall results for the estimation and mapping of forest height, and its conversion into volume and above-ground biomass (AGB). The processing chain was first designed for small dataset, this study shows that results are interesting on both a local and national scale using GEDI height metrics as reference data. Optical and radar combination with spatial texture indices, used with a simple machine learning algorithm, allows accurate estimations of forest parameter both on a local and national scale.

Forest height, volume and AGB maps are available on Zenodo. The maps are from 2020, at 10 m spatial resolution, and the Copernicus Dominent Leaf Type (DLT, HRlayers) mask was applied to mask non-forest areas. Link to the maps: https://zenodo.org/doi/10.5281/zenodo.8071003.

The maps can be used on a local scale, the validation on the NFI plots and on the Lidar CHMs shows absolute errors around 20% for the height, 40-45% for the volume and AGB. These maps can help to implement forest management plans or carbon balances on a territorial scale.

A saturation effect persists in these estimations, doubled for AGB and volume because the power-law relationships between forest height and AGB or volume accentuate the saturation. The saturation comes partly from the fact that the satellite signal saturates for high values. It can also come in part from the distribution of the input data: Gaussian distributions with high concentrations of average values, making it difficult for simple algorithms to accurately estimate low and high values. The use of Linear Forest Regression, instead of a simple Random Forest regression, already mitigate this issue. Deep learning approaches could partly resolve the saturation problem, but they also make the transmission of the method more complicated and make difficult the interpretation of the physics behind the relationships between satellite signal and forest parameters.

## Acknowledgments

We would like to thank the "Agence de la transition écologique" (ADEME) for funding this work, the "Office National des Forêts" (ONF) for providing us with the ALS data and measurement dataset, and the "Laboratoire de l'Inventaire Forestier" (LIF, IGN) for validating the height map with the inventory plots.